# Foundation Models for Slide-level Cancer Subtyping in Digital Pathology

Pablo Meseguer[1,2][0000-0001-7821-6168], Rocío del Amor[1][0000-0002-5342-2093], Adrian Colomer[1][0000-0002-7616-6029], and Valery Naranjo[1,2][0000-0002-7616-6029]

[1] Instituto Universitario de Investigación en Tecnología Centrada en el Ser Humano (HUMAN-Tech), Universitat Politécnica de Valencia, 46022, Spain
[2] Valencian Graduate School and Research Network for Artificial Intelligence (valgraAI), Valencia, 46022, Spain
{pabmees, madeam2, adcogra, vnaranjo}@upv.es

**Abstract.** Since the emergence of the ImageNet dataset, the pretraining and fine-tuning approach has become widely adopted in computer vision due to the ability of ImageNet-pretrained models to learn a wide variety of visual features. However, a significant challenge arises when adapting these models to domain-specific fields, such as digital pathology, due to substantial gaps between domains. To address this limitation, foundation models (FM) have been trained on large-scale in-domain datasets to learn the intricate features of histopathology images. In cancer diagnosis, whole-slide image (WSI) prediction is essential for patient prognosis, and multiple instance learning (MIL) has been implemented to handle the giga-pixel size of WSI. As MIL frameworks rely on patch-level feature aggregation, this work aims to compare the performance of various feature extractors developed under different pretraining strategies for cancer subtyping on WSI under a MIL framework. Results demonstrate the ability of foundation models to surpass ImageNet-pretrained models for the prediction of six skin cancer subtypes.

## 1  Introduction

The implementation of transfer learning (TL) methodologies, which leverage the extensive visual knowledge encoded in ImageNet-pretrained models [12], has become the most adopted strategy in computer vision (CV). TL-based approaches have gained significant attention due to the ability of pre-trained models to facilitate fine-tuning of diverse downstream tasks where data availability is scarce. However, the adaptation of such models to domain-specific challenges is hindered due to significant domain gaps [8]. This limitation is especially significant in medical imaging in general and computational pathology in particular, requiring the development of models that can learn the unique characteristics of these images.

Aiming to obtain discriminative features tailored to histopathological images, different learning paradigms that go beyond the fully supervised approach of ImageNet pretraining have been proposed to develop in-domain foundation models. Self-supervised learning (SSL) pretends to learn visual representations using the images themselves, without any other form of supervision.. In particular, SSL techniques aim to learn feature representations from histopathological images under contrastive learning frameworks, such as SimCLR [4] and convolutional neural networks (CNN) [13] or histopathology-oriented approaches using Transformers architectures [19]. Recently, contrastive language image pretraining (CLIP) has been proposed to learn vision features from vision language supervision (VLS) and enable zero-shot transfer [17]. To accommodate CLIP models to histopathological tasks, large-scale datasets of histopathology image-caption have been gathered from diverse sources, such as YouTube, [10], Twitter [9] or PubMed articles [15].

In recent years, computational pathology has permitted digitizing tissue sections into whole-slide images (WSI), promoting the application of CV algorithms [1]. The giga-pixel size of WSI and the time and resource-consuming task of pixel-level annotation have motivated the implementation of weakly-supervised frameworks based on multiple instance learning (MIL) [2]. This approach aims to train models that can predict the global label of the slide rather than the label of each instance, thus replicating the diagnostic process of pathologists. In particular, we focus on cutaneous spindle cell (CSC) neoplasms for cancer subtyping, a family of skin tumors characterized by spindle-shaped neoplastic cells with varying degrees of differentiation [20]. This group of neoplasms poses a challenge in the diagnosis due to notable morphological overlap between subtypes, suggesting that implementing DL-based algorithms could contribute to addressing discrepancies between pathologists [7].

Previous works introducing novel pretraining strategies assess only their performance compared to models following similar frameworks. Specifically, [15] compared four VLS strategies, while [3] did so with SSL techniques. We lack in-depth comparative analysis across pretraining strategies for slide-level predictions under MIL frameworks. This work aims to address this gap by investigating the performance of ImageNet-pretrained against in-domain FM trained under self-supervised and vision-language supervised paradigms for skin cancer subtyping at the slide level. By conducting a thorough comparative analysis, we aim to provide insights into the effectiveness of different pretraining strategies and MIL aggregations in enhancing the accuracy and robustness of skin cancer subtyping models.

## 2    Methodology

### 2.1    Problem formulation

In MIL frameworks, WSIs are arranged in bags $X$ containing an arbitrary number of instances $N$, such as $X = \{x_n\}_{n=1}^{N}$. Each bag is assigned to one of $S$ mutually exclusive classes in the multi-class scenario, denoted as $Y_s \in \{0, 1\}$.

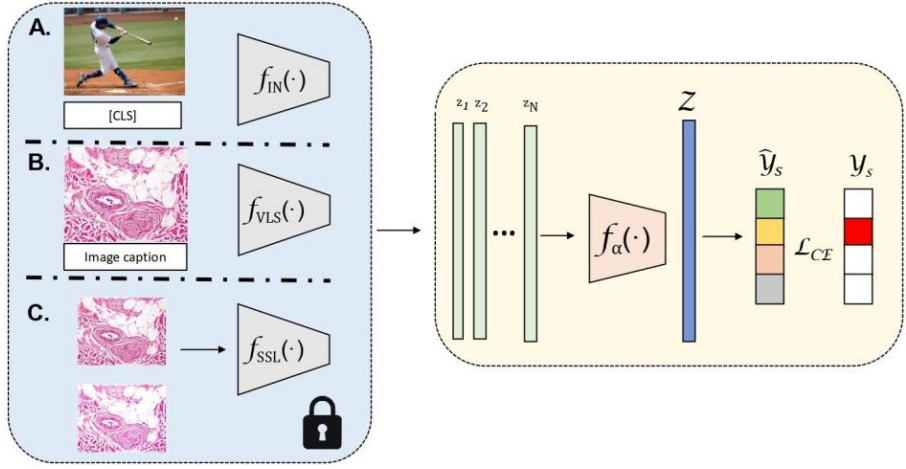

**Fig. 1.** Method overview. Comparative analysis of different pretraining strategies for slide-level prediction of cancer subtypes: ImageNet pretraining (A.), vision-language supervision (B.) and self-supervised learning (C.).

In the MIL embedding formulation, each instance does not necessarily require an associated label, and the aim is to predict the slide-level label by levering features extracted at the patch level.

The proposed framework for evaluating different pretraining paradigms for cancer subtyping at the slide level is presented in Figure 1. We introduce a feature extractor, $f_\vartheta( ) : X \to F$, where $\vartheta$ stands for the parameters of each encoder trained under different learning paradigms. $f_\vartheta$ projects instances $x \in X$ to a lower-dimensional manifold $z \in F \subset R^d$, where $d$ is the embedding dimension. Note that $\vartheta$ remains frozen during the training phase as we work directly on top of the instance-level projections extracted by the different backbones, thus reducing the computational cost of fine-tuning and promoting model efficiency. This approach ensures a controlled evaluation of foundation model pretraining paradigms under a multiple instance paradigm for cancer subtyping in WSI.

Subsequently, we define a MIL aggregation function, $f_\alpha( ) : F \to Z$, which combines the instance-level projections into a global embedding ($Z$). MIL aggregations should be independent of the number of instances in each bag and different functions have been investigated in the Section 4. Finally, a multilayer perceptron classifier, $f_\phi( ) : Z \to \hat{Y}_s$, predicts softmax bag-level class scores ($\hat{Y}_s$), where $\hat{Y}_s \in [0, 1]$. The optimization of the model parameters $\alpha$, and $\phi$ is guided by minimizing the standard categorical CE loss between the reference labels and predicted scores, as follows:

$$L_{CE} = -\frac{1}{S}\sum_{s=1}^{S} Y_s log(\hat{Y}_s) \qquad (1)$$

## 2.2   Instance-level feature extraction

Despite the notable domain gap between natural and medical images, models pretrained on ImageNet have been widely adopted to carry out tasks with histopathology images [6,14,5]. However, the emergence of foundation models trained on in-domain large scale datasets promises to obtain more rich and discriminative features on histopathology images that will enhance downstream task adaptation. For that purpose, this work compares the feature extractions ability of three pretraining strategies to obtain patch-level representations used as input of MIL frameworks.

While $f_{IN}(\cdot)$ denotes a CNN backbone pretrained on the ImageNet dataset using fully supervised learning (see Figure 1 A.), $f_{VLS}(\cdot)$ and $f_{SSL}(\cdot)$ explore the potential of FM trained with in-domain data. $f_{IN}(\cdot)$ is likely to have learned general features from natural images that could not transfer correctly to other domains. Specifically, $f_{VLS}(\cdot)$ (see Figure 1 B.) incorporates both an image and text encoder, which project paired image-caption histopathological data onto a lower-dimensional manifold. It then employs contrastive learning to learn visual features by aligning both latent spaces. On the other hand, self-supervised learning $f_{SSL}(\cdot)$ (see Figure 1 C.) also utilizes contrastive learning but relies on various views of the same histopathology image, which are obtained through data augmentation to understand the intrinsic properties and variations within the image.

## 3   Experimental setting

In this section, we present a comprehensive descriptions of the dataset used in this study and an overview of the experimental framework proposed for the paradigm comparison.

### 3.1   Materials

For the evaluation of the proposed framework, we resort to a dataset containing up to 608 histopathology slides of six subtypes of skin cancer. Details about the number of images for each calss are provided at Table 1). In particular, cutaneous spindle cell neoplasms: leiomyoma (lm), leiomyosarcoma (lms), dermatofibroma (df), dermatofibrosarcoma (dfs), spindle cell melanoma (mfc) and fibroxanthoma (fxa). Each sample was annotated at the slide-level by two expert pathologist [7]. Note that the dataset contains images from two hospitals: Hospital Clínico of Valencia (HCUV) and the Hospital Universitario San Cecilio (HUSC) of Granada.

Following previous work using this dataset [7], the WSI of the dataset followed these pre-processing steps. Initially, the slides where downsampled to 10x magnification and the tissue was separated from the background using Otsu thresholding for segmentation. To accommodate the MIL paradigm, each WSI is cropped using an in-house software into patches of 512 squared pixels with a 50% overlap. To reduce the noise of the input data, patches with more than 20% of background were marked as non-informative and discarded to create the bags.

**Table 1.** Description of the dataset used containing WSI of skin cancer subtypes from two centers: Valencia (HCUV) and Granada (HUSC).

|  | **HCUV** | **HUSC** | **Total** |
|---|---|---|---|
| **Leiomyoma** | 31 | 71 | 102 |
| **Leiomyosarcoma** | 25 | 26 | 51 |
| **Dermatofibroma** | 73 | 95 | 168 |
| **Dermatofibrosarcoma** | 17 | 44 | 61 |
| **Melanoma** | 49 | 73 | 122 |
| **Fibroxanthoma** | 44 | 60 | 104 |
| **All classes** | 239 | 369 | 608 |

### 3.2 Experimental framework

We follow a 5-fold stratified cross-validation, and all models are evaluated in terms of balanced accuracy to consider unbalanced classes equally. Following [15], models are trained for 20 epochs with an AdamW optimizer, a cosine learning rate scheduler, and a peak learning rate of 1e-4. All models have been implemented Python 3.10 and PyTorch 1.12.1.

We selected one representative method for each framework to assess the comparison of different pretraining paradigms. For the ImageNet-pretrained model, we used VGG16 as CNN backbone as it showed superior performance to other feature extractors in similar tasks [5]. Among SSL paradigms, TransPath [19] incorporates a Swin Transformer to a CNN architecture and develops a method oriented for histopathology images named semantically-relevant contrastive learning (SRCL). In particular, it incorporates random cropping, color distorsions an Gaussian blur as data augmentation techniques following SimCLR [4]. Among vision-languages models, Pathology Language Image Pretraining (PLIP) [9] gathered a large dataset of pathology images with paired image descriptions from Twitter. It was used to fine-tune CLIP-like models for diverse downstream tasks that exploit the multimodal ability, such as zero-shot transfer and cross-model retrieval. We followed the original implementation of each model for feature extraction, thus leading to a larger embedding dimension for the SSL model ($d = 758$) than the IN and VLS models ($d = 512$).

## 4   Results

Here, a series of quantitative and qualitative analysis of the results and the date is presented to demonstrate the contribution of the framework to solve the task at hand.

**Table 2.** Quantitative results in terms of balanced accuracy for the different MIL aggregations and pretraining paradigms: ImageNet pretraining (IN), vision-language supervision (VLS) and self supervised learning (SSL).

|                      | IN     | VLS        | SSL        |
|----------------------|--------|------------|------------|
| **MI-SimpleShot [15]** | 0.5068 | **0.6498** | 0.5779     |
| **BGAP**             | 0.6406 | 0.7109     | **0.7586** |
| **ABMIL [11]**       | 0.6287 | 0.7385     | **0.7397** |
| **TransMIL [18]**    | 0.5713 | **0.7974** | 0.7206     |

## 4.1   Quantitative analysis

The evaluation of the different pretraining strategies in the context of slide-level cancer subtyping is presented in Table 2. For that purpose, we evaluate the three paradigms with different MIL aggregation functions such as batch global average pooling (BGAP), attention-based MIL (ABMIL) [11] to find the most relevant instances, and transformer-based MIL (TransMIL) [18] to assess the correlation between instances. Thus, we leverage the ability of the MIL aggregations to find the most critical samples and their relationship to address the classification problem. We also compare the MI-SimpleShot [15], a non-trainable classification based on representational learning that constructs the class prototype by averaging the instance-level vectors of all samples within a class.

Results show that ImageNet pretraining underperforms by a large margin (from 11.10% to 22.61%) compared to the other feature extractors across the different classification methods. Its performance degradation is notable under the MI-SimpleShot method based on representational learning, highlighting that features extracted by the IN model do not transfer well for histopathology images. When assessing the performance of foundation models trained on in-domain datasets, the PLIP vision language model (VLM) shows improved performance in representational learning, achieving 64.98% balanced accuracy in a distance-based classification method, such as MI-SimpleShot. Across the different aggregation methods for the VLS model, TransMIL outperforms BGAP and ABMIL. Moreover, figures of merit for the SSL could suggest that TransMIL aggregation leads to overfitting as it contains 40x more trainable parameters than ABMIL. This issue is aggravated due to the larger dimension of the latent space of the SSL compared to the vision language model.

## 4.2   Qualitative analysis

To qualitative evaluate the feature extraction ability of each pretraining paradigm, we plot the two-dimensional T-SNE [16] in Figure 2. For that purpose, we performed the BGAP aggregation of all the instance-level embedding within a WSI to obtain the global embedding representing the slide. With this approach, we strictly evaluate the feature extraction ability of each paradigm as trainable functions are not incorporated.

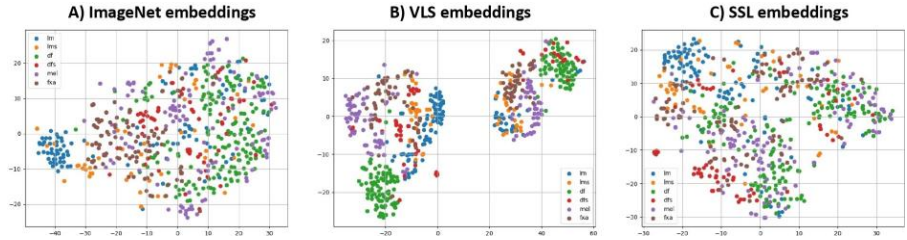

**Fig. 2.** TSNE visualization of the batch global average (BGAP) embeddings of each feature extractor trained with different pretraining paradigms: ImageNet (IN), vision language supervision (VLS) and self-supervised learning (SSL).

The qualitative visualization of the features extracted by the vision-language model shows a notable bias between the images of the two hospitals, as two main clusters are created after dimensionality reduction. However, PLIP's better representational learning compared to other models is shown when analyzing each separated cluster as the samples of each class are closer to each other. Despite the inherent bias to the stain in these embeddings, MIL models manage to handle this variability, achieving a balanced accuracy larger than 70%. The representation of SSL embeddings shows a reduction in stain bias and discriminative ability between classes.

## 5    Conclusion

Foundation models hold great promise in solving many downstream tasks due to the training with large-scale datasets. Moreover, domain gaps between natural and medical images have promoted the development of FM to address particular domains, such as computation pathology. In this field, the explored pretrained strategies work at the region or patch level due to the limitation inherent to the enormous size of WSI that can span up to 10 thousand squared pixels. However, this work has demonstrated that the instance-level feature extraction of the FM can be extrapolated to construct MIL paradigms that allow the prediction at the slide level, which mimics the decision making of pathologist thus promoting their application in real clinical settings. When comparing different pretraining strategies, vision-language supervision shows greater representation learning than self-supervision paradigms, although the incorporation of MIL aggregation proves to achieve comparable performance for both methodologies.

The main limitation of this work is that the comparison across different pretraining paradigms is restricted to one model within each strategy. Additionally, the framework is validated in a multi-center dataset with WSI with two different centers, which supposes a limitation, especially when using the image encoder of the VLM as the instance-level encoder. However, this handicap is not manageable with stain normalization as each tumor subtype within the dataset is composed of cells that react differently to the eosin and hematoxylin stain. Future lines could investigate novel color normalization methods that work on top of foundation models that could have a more significant bias to the stains and scanners of different centers.

**Funding**: This work has received funding from the Spanish Ministry of Economy and Competitiveness through projects PID2019-105142RB-C21 (AI4SKIN) and PID2022-140189OB-C21 (ASSIST). The work of Rocío del Amor and Pablo Meseguer has been supported by the Spanish Ministry of Universities under an FPU Grant (FPU20/05263) and valgrAI - Valencian Graduate School and Research Network of Artificial Intelligence, respectively.

## References

1. Abels, E., Pantanowitz, L., Aeffner, F., Zarella, M.D., van der Laak, J., Bui, M.M., Vemuri, V.N., Parwani, A.V., Gibbs, J., Agosto-Arroyo, E., Beck, A.H., Kozlowski, C.: Computational pathology definitions, best practices, and recommendations for regulatory guidance: a white paper from the digital pathology association. The Journal of Pathology **249**(3), 286–294 (2019). https://doi.org/https://doi.org/10.1002/path.5331

2. Campanella, G., Hanna, M.G., Geneslaw, L., Miraflor, A., Werneck Krauss Silva, V., Busam, K.J., Brogi, E., Reuter, V.E., Klimstra, D.S., Fuchs, T.J.: Clinical-grade computational pathology using weakly supervised deep learning on whole slide images. Nature Medicine **25**(8), 1301–1309 (08 2019). https://doi.org/10.1038/s41591-019-0508-1

3. Chen, R.J., Ding, T., Lu, M.Y., Williamson, D.F.K., Jaume, G., Song, A.H., Chen, B., Zhang, A., Shao, D., Shaban, M., Williams, M., Oldenburg, L., Weishaupt, L.L., Wang, J.J., Vaidya, A., Le, L.P., Gerber, G., Sahai, S., Williams, W., Mahmood, F.: Towards a general-purpose foundation model for computational pathology. Nature Medicine **30**(3), 850–862 (03 2024). https://doi.org/10.1038/s41591-024-02857-3

4. Chen, T., Kornblith, S., Norouzi, M., Hinton, G.: A simple framework for contrastive learning of visual representations. In: III, H.D., Singh, A. (eds.) Proceedings of the 37th International Conference on Machine Learning. Proceedings of Machine Learning Research, vol. 119, pp. 1597–1607. PMLR (13–18 Jul 2020), https://proceedings.mlr.press/v119/chen20j.html

5. del Amor, R., Launet, L., Colomer, A., Moscardó, A., Mosquera-Zamudio, A., Monteagudo, C., Naranjo, V.: An attention-based weakly supervised framework for spitzoid melanocytic lesion diagnosis in whole slide images. Artificial Intelligence in Medicine **121**, 102197 (2021). https://doi.org/https://doi.org/10.1016/j.artmed.2021.102197

6. Del Amor, R., Meseguer, P., Parigi, T.L., Villanacci, V., Colomer, A., Launet, L., Bazarova, A., Tontini, G.E., Bisschops, R., de Hertogh, G., Ferraz, J.G., Götz, M., Gui, X., Hayee, B., Lazarev, M., Panaccione, R., Parra-Blanco, A., Bhandari, P., Pastorelli, L., Rath, T., Røyset, E.S., Vieth, M., Zardo, D., Grisan, E., Ghosh, S., Iacucci, M., Naranjo, V.: Constrained multiple instance learning for ulcerative colitis prediction using histological images. Computer Methods and Programs in Biomedicine **224**, 107012 (2022). https://doi.org/10.1016/j.cmpb.2022.107012

7. del Amor, R., Pérez-Cano, J., López-Pérez, M., Terradez, L., Aneiros-Fernandez, J., Morales, S., Mateos, J., Molina, R., Naranjo, V.: Annotation protocol and crowdsourcing multiple instance learning classification of skin histological images: The cr-ai4skin dataset. Artificial Intelligence in Medicine **145**, 102686 (2023). https://doi.org/https://doi.org/10.1016/j.artmed.2023.102686

8. Guan, H., Liu, M.: Domain adaptation for medical image analysis: A survey. IEEE Transactions on Biomedical Engineering **69**(3), 1173–1185 (2022). https://doi.org/10.1109/TBME.2021.3117407

9. Huang, Z., Bianchi, F., Yuksekgonul, M., Montine, T.J., Zou, J.: A visual–language foundation model for pathology image analysis using medical twitter. Nature Medicine **29**(9), 2307–2316 (09 2023). https://doi.org/10.1038/s41591-023-02504-3

10. Ikezogwo, W., Seyfioglu, S., Ghezloo, F., Geva, D., Sheikh Mohammed, F., Anand, P.K., Krishna, R., Shapiro, L.: Quilt-1m: One million image-text pairs for histopathology. In: Oh, A., Neumann, T., Globerson, A., Saenko, K., Hardt, M., Levine, S. (eds.) Advances in Neural Information Processing Systems. vol. 36, pp. 37995–38017. Curran Associates, Inc. (2023), https://proceedings.neurips.cc/paper_files/paper/2023/file/775ec578876fa6812c062644964b9870-Paper-Datasets_and_Benchmarks.pdf

11. Ilse, M., Tomczak, J., Welling, M.: Attention-based deep multiple instance learning. In: Dy, J., Krause, A. (eds.) Proceedings of the 35th International Conference on Machine Learning. Proceedings of Machine Learning Research, vol. 80, pp. 2127–2136. PMLR (10–15 Jul 2018), https://proceedings.mlr.press/v80/ilse18a.html

12. Krizhevsky, A., Sutskever, I., Hinton, G.E.: Imagenet classification with deep convolutional neural networks. Commun. ACM **60**(6), 84–90 (may 2017). https://doi.org/10.1145/3065386

13. Li, B., Li, Y., Eliceiri, K.W.: Dual-stream multiple instance learning network for whole slide image classification with self-supervised contrastive learning. In: Proceedings of the IEEE/CVF Conference on Computer Vision and Pattern Recognition (CVPR). pp. 14318–14328 (June 2021)

14. Liu, M., Liu, Y., Cui, H., Li, C., Ma, J.: Mgct: Mutual-guided cross-modality transformer for survival outcome prediction using integrative histopathology-genomic features. In: 2023 IEEE International Conference on Bioinformatics and Biomedicine (BIBM). pp. 1306–1312 (2023). https://doi.org/10.1109/BIBM58861.2023.10385897

15. Lu, M.Y., Chen, B., Williamson, D.F.K., Chen, R.J., Liang, I., Ding, T., Jaume, G., Odintsov, I., Le, L.P., Gerber, G., Parwani, A.V., Zhang, A., Mahmood, F.: A visual-language foundation model for computational pathology. Nature Medicine **30**(3), 863–874 (2024). https://doi.org/10.1038/s41591-024-02856-4

16. van der Maaten, L., Hinton, G.: Visualizing data using t-sne. Journal of Machine Learning Research **9**(86), 2579–2605 (2008), http://jmlr.org/papers/v9/vandermaaten08a.html

17. Radford, A., Kim, J.W., Hallacy, C., Ramesh, A., Goh, G., Agarwal, S., Sastry, G., Askell, A., Mishkin, P., Clark, J., Krueger, G., Sutskever, I.: Learning transferable visual models from natural language supervision. In: Meila, M., Zhang, T. (eds.) Proceedings of the 38th International Conference on Machine Learning. Proceedings of Machine Learning Research, vol. 139, pp. 8748–8763. PMLR (18–24 Jul 2021), https://proceedings.mlr.press/v139/radford21a.html

18. Shao, Z., Bian, H., Chen, Y., Wang, Y., Zhang, J., Ji, X., zhang, y.: Transmil: Transformer based correlated multiple instance learning for whole slide image classification. In: Ranzato, M., Beygelzimer, A., Dauphin, Y., Liang, P., Vaughan, J.W. (eds.) Advances in Neural Information Processing Systems. vol. 34, pp. 2136–2147. Curran Associates, Inc. (2021), https://proceedings.neurips.cc/paper_files/paper/2021/file/10c272d06794d3e5785d5e7c5356e9ff-Paper.pdf

19. Wang, X., Yang, S., Zhang, J., Wang, M., Zhang, J., Yang, W., Huang, J., Han, X.: Transformer-based unsupervised contrastive learning for histopathological image classification. Medical Image Analysis **81**, 102559 (2022). https://doi.org/10.1016/j.media.2022.102559
20. Winnepenninckx, V., De Vos, R., Stas, M., van den Oord, J.J.: New phenotypical and ultrastructural findings in spindle cell (desmoplastic/neurotropic) melanoma. Applied Immunohistochemistry & Molecular Morphology **11**(4), 369–375 (2003). https://doi.org/10.1097/01.PAI.0000040947.01212.40